\documentclass[conference]{IEEEtran}
\IEEEoverridecommandlockouts

\usepackage{setspace}

\usepackage{amsmath,amssymb,amsfonts}
\usepackage{algorithmic}
\usepackage{graphicx}
\usepackage{textcomp}
\usepackage{xcolor}
\usepackage{float}

\usepackage{booktabs}
\usepackage{pifont}  
\newcommand{\cmark}{\ding{51}}
\newcommand{\xmark}{\ding{55}}
\usepackage{array}
\usepackage{multirow}

\usepackage{hyperref}
\hypersetup{
    colorlinks=false,        
    linkbordercolor=blue,    
    citebordercolor=blue,
    urlbordercolor=blue,
    pdfborder={0 0 1}       
}

\def\BibTeX{{\rm B\kern-.05em{\sc i\kern-.025em b}\kern-.08em
    T\kern-.1667em\lower.7ex\hbox{E}\kern-.125emX}}

\usepackage[backend=biber,style=ieee]{biblatex}
\begin{document}

\title{Understanding Fault Tolerance of Adversarially Robust Pruned Models\\
}

\author{\IEEEauthorblockN{Manali Dangarikar and Cory Merkel}
\IEEEauthorblockA{\{md2591, cemeec\}@rit.edu}
{Brain Lab} \\
{Rochester Institute of Technology}\\
Rochester, NY, USA

}

\maketitle

\begin{abstract}
Deep neural networks (DNNs) deployed on resource-constrained neuromorphic hardware face three concurrent challenges: the need for model compression through pruning, vulnerability to adversarial input perturbations, and susceptibility to hardware-induced weight faults such as stuck-at-zero errors. While each of these factors has been studied in isolation, their combined effects on model reliability have received little attention. This paper presents an empirical investigation of how pruning, adversarial training, and hardware fault injection interact to affect the robustness of convolutional neural networks. Using a compact three-layer CNN trained on MNIST, we conduct three experiments: (1) comparing the fault tolerance of naturally and adversarially trained models under simultaneous hardware faults and adversarial attacks, (2) evaluating how pruning affects adversarial robustness, and (3) characterizing the joint accuracy surface across fault rates, adversarial perturbation magnitudes, and pruning levels. Our results show that adversarial training improves robustness against input perturbations but increases sensitivity to stuck-at-zero weight faults. Contrary to intuition, pruning did not significantly increase fault sensitivity, and varying the pruning level had little effect across fault rates and attack strengths. These results highlight the need to jointly consider adversarial robustness and hardware reliability.
\end{abstract}

\section{Introduction}
The growing demand for on-device inference has driven the deployment of deep neural networks (DNNs) onto edge computing platforms, including neuromorphic hardware such as IBM NorthPole and BrainChip Akida. However, operating in these environments places three simultaneous demands on a model: it must be compact enough to fit within tight memory and compute budgets \cite{lecun1989optimal, han2015learning, han2015deep}, robust enough to resist adversarial input perturbations \cite{goodfellow2014explaining, madry2017towards}, and tolerant of hardware-induced weight faults \cite{hong2019terminal, weng2020towards}. Each of these requirements has been studied extensively in isolation, yet their combined effect on model reliability has received little attention.

This work investigates how model compression through pruning, adversarial training, and stuck-at weight faults jointly affect the reliability of convolutional neural networks. Specifically, we study how adversarially trained models compare to naturally trained models under weight faults, and the effect of pruning in adversarially trained networks on both input perturbations and weight fault tolerance.

Model pruning is one of the most effective techniques for reducing the computational cost of DNNs. It removes parameters that contribute little to the model's output, resulting in a smaller and faster network suitable for edge deployments \cite{lecun1989optimal, han2015learning, han2015deep}. Pruning strategies have been widely studied and can be broadly categorized by how pruning is performed, when it is applied during training, and the criteria used to identify parameters for removal \cite{cheng2024survey}.

When implemented on neuromorphic chips, DNNs are exposed to hardware-level faults that can alter their learned behavior. A common example is the stuck-at fault, where a synaptic weight becomes permanently fixed at either logic low (0) or logic high (1), regardless of input. These faults can arise from manufacturing defects, device degradation, or environmental stress, and can also be deliberately induced by an attacker \cite{hong2019terminal}. They distort the internal computations of the network, degrading inference reliability \cite{segee1991fault, hsieh2021fault}. Prior work has shown that convolutional networks exhibit layer-dependent sensitivity to random weight perturbations, with lower layers being substantially more fragile than higher layers \cite{cheney2017robustness}. As a result, fault tolerance has become a critical concern for systems deployed on neuromorphic hardware.

In this work, we focus on stuck-at-zero faults, which are simulated through software-based fault injection on a floating-point model. This approach allows controlled study of fault effects but does not capture the full behavior of real neuromorphic hardware, such as analog noise or device-specific fault patterns. We use stuck-at-zero injection as a proxy for hardware-level weight corruption, consistent with prior simulation-based fault studies \cite{cheney2017robustness}.

Beyond their individual effects, pruning and hardware faults can each alter the network's decision boundaries, influencing its response to adversarial examples. Adversarial examples are slightly perturbed inputs that cause high-confidence misclassification. These perturbations are often imperceptible to humans but can drastically alter a model's output \cite{goodfellow2014explaining}, \cite{madry2017towards}. Adversarial training serves as the most common defense technique against input perturbations and is achieved by augmenting training data with adversarially perturbed examples.

Prior studies have reported that pruning can improve adversarial robustness within certain compression limits, but beyond a critical pruning rate, the model becomes more vulnerable to adversarial attacks \cite{pavlitska2023relationship, wang2018adversarial, jordao2021effect, guo2018sparse}. How this relationship is further affected when the network is also subjected to hardware faults remains an open question.

To the best of our knowledge, no prior work has jointly studied the interaction of pruning, adversarial training, and hardware fault tolerance in convolutional networks. While \cite{tsai2021non} demonstrated that adversarial training increases sensitivity to weight perturbations in fully-connected networks, their work did not examine convolutional architectures, model compression, or neuromorphic-specific fault models, as summarized in Table~\ref{tab:prior_work}.

\begin{table}[ht]
\centering
\caption{Summary of prior work addressing pruning, adversarial robustness, and hardware fault tolerance. This work is the first to jointly address all three.}
\label{tab:prior_work}
\resizebox{\columnwidth}{!}{%
\renewcommand{\arraystretch}{1.3}
\begin{tabular}{lccc}
\toprule
\textbf{Representative Works} & \textbf{Pruning} & \textbf{Adversarial Robustness} & \textbf{Weight Faults} \\
\midrule
\cite{lecun1989optimal, han2015learning, han2015deep}
    & \cmark & \xmark & \xmark         \\
\cite{goodfellow2014explaining, madry2017towards}
    & \xmark & \cmark & \xmark  \\
\cite{cheney2017robustness, hong2019terminal, hsieh2021fault}
    & \xmark & \xmark & \cmark \\
\cite{wang2018adversarial, guo2018sparse, ye2019adversarial, jordao2021effect, pavlitska2023relationship}
    & \cmark & \cmark & \xmark  \\
\cite{weng2020towards, tsai2021non}
    & \xmark & \cmark & \cmark \\
\midrule
\textbf{This work}  & \cmark & \cmark & \cmark \\
\bottomrule
\end{tabular}
}
\end{table}

This paper bridges this gap through controlled experiments that jointly vary pruning level, adversarial training, and fault injection rate. Our contributions are:
\begin{enumerate}
    \item We provide joint characterization of how adversarial training, model compression, and hardware faults interact to determine model reliability.
    
    \item We reveal non-obvious interactions between these threats: adversarially trained models exhibit heightened sensitivity to hardware faults.

    \item Intuitively we would expect compressed networks to be more prone to weight faults since they have fewer weights to represent information. However, we observe that model compression has minimal impact on adversarial robustness in fault-free settings. 
\end{enumerate}

\section{Related Work}

\subsection{Adversarial Robustness}
Adversarial examples were first defined by \cite{szegedy2013intriguing}, revealing that imperceptible input perturbations can cause confident misclassifications in deep neural networks. This discovery led to the development of gradient-based attacks such as the Fast Gradient Sign Method (FGSM) \cite{goodfellow2014explaining} and Projected Gradient Descent (PGD) \cite{madry2017towards}, which have become standard benchmarks for evaluating model vulnerability.
Adversarial training, which augments the training set with adversarially perturbed examples, has emerged as the most effective defense mechanism \cite{madry2017towards}. Although this approach improves robustness to adversarial perturbations, it often degrades clean accuracy and substantially increases training cost \cite{ye2019adversarial}.

\subsection{Model Compression and Pruning}
The motivation for pruning comes from the need to reduce the energy required to run large networks so that they can run in real time on edge devices. Early pruning methods such as \cite{lecun1989optimal} used Hessian information to calculate the sensitivity, which in turn decided the parameters to be removed. Another early work by \cite{segee1991fault} show that there is a strong correlation between node relevance and weight magnitude. Moreover, they also showed that pruning does not have a significant impact on the tolerance of the network subject to zeroing of a single weight. Later work demonstrated that simpler threshold-based pruning with iterative retraining could achieve 9$\times$ to 13$\times$ compression on AlexNet and VGGNet respectively \cite{han2015learning}. The compression pipeline in \cite{han2015deep} combined pruning, quantization with weight sharing, and Huffman coding to achieve 35$\times$ compression on AlexNet. Structured pruning methods such as \cite{li2017pruning} remove entire filters or channels for hardware efficiency, whereas unstructured pruning achieves higher sparsity at the cost of irregular memory access. Recent survey \cite{cheng2024survey} categorizes pruning techniques by granularity (unstructured \textit{vs.} structured), timing (pre-training, during-training, or post-training), and selection criteria (magnitude, gradient, or saliency-based).

\subsection{Hardware Fault Tolerance}
Hardware faults pose a reliability challenge for neural networks deployed on neuromorphic and edge devices. \cite{cheney2017robustness} demonstrated that convolutional networks exhibit layer-dependent sensitivity to weight perturbations, with lower layers more fragile than higher layers. \cite{hong2019terminal} showed that DNNs are vulnerable to bit-flip corruptions. Their analysis shows that a Rowhammer-based attacker can cause upto 99\% drop in accuracy with constrained bit-flip corruption and with no knowledge of the model. Unlike deliberate fault-injection attacks that adversarially craft bit-flips to target specific weights \cite{hong2019terminal, rakin_tbt_2020}, this work focuses on random hardware-reliability faults. \cite{weng2020towards} provides a perturbation region such that DNNs will maintain their accuracy if weight perturbations are within that region. Further, they provide a perturbation-aware weight quantization technique showing significant improvement. \cite{hsieh2021fault} talk about different fault patterns in neuromorphic hardware and provide ways to detect them. The permanent stuck-at faults (where weights become fixed at zero or one) represent common failure modes in memristive hardware causing the need for fault-aware design.

\subsection{Pruning and Adversarial Robustness}
The interaction between model compression and adversarial robustness has produced contradictory findings across studies. \cite{wang2018adversarial} used magnitude-based pruning and found that while it maintains clean accuracy, heavy pruning substantially reduces adversarial robustness. In contrast, \cite{jordao2021effect} used structured pruning and found that pruning alone, under standard natural training, improves adversarial robustness by acting as a regularizer, achieving results competitive with adversarial training without requiring it. \cite{guo2018sparse} similarly report that increased sparsity improves robustness. This disagreement suggests that the pruning granularity, timing, criteria, as well as the training type - natural vs. adversarial, shapes whether pruning helps or hurts adversarial robustness. These differences motivate our use of magnitude-based (unstructured) pruning combined with adversarial training as a distinct setting from prior work.

\subsection{Adversarial Robustness and Hardware Faults}
Recent work has explored the interaction between adversarial robustness and weight perturbations. \cite{tsai2021non} showed that adversarially trained models are more vulnerable to weight perturbations than naturally trained models and proposed a margin-based regularization method to improve robustness under joint perturbations. However, their work focused on dense networks and did not consider model compression or neuromorphic fault models such as stuck-at-zero faults. Our work extends this line of research by jointly characterizing the effects of pruning, adversarial training, and hardware fault tolerance in convolutional neural networks.

\section{Method}
This study extends the framework of \cite{wang2018adversarial} on the adversarial robustness of pruned neural networks to ensure consistency. A compact CNN is trained on MNIST, subjected to pruning and adversarial training, and evaluated under simulated hardware faults. Both clean and adversarial accuracies are measured across varying attack strengths, pruning levels, and weight faults. All experiments are conducted in PyTorch.

\subsection{Network Architecture and Training}
The network consists of three convolutional layers with 32, 64, and 64 filters, followed by a fully connected layer of 10 neurons before softmax. The convolutional layers have a kernel size/ stride/ padding of 3/ 1/ 1 respectively. Natural training uses cross-entropy loss with RMSProp optimizer (learning rate = 0.001, weight decay = 5e-4) for 10 epochs. In case of adversarial training, another 5 epochs with natural as well as adversarial images are carried out. Each layer has a binary mask \(M\) applied directly to its weight tensor \(W\). The mask is stored as a non-trainable variable and multiplied element-wise with the corresponding weights to deactivate selected connections:
\begin{equation}
    W = W \odot M
\end{equation}

\subsection{Pruning Procedure}
After baseline training, X\% of weights with the smallest absolute values are set to zero globally. We experimented with 20, 40, 60 and 80 percent of pruning. The resulting pruned network is retrained using the same hyper-parameters to recover lost accuracy and stabilize performance. Three model variants are obtained:
(1) Naturally Trained,
(2) Adversarially Trained,
(3) Adversarially Trained and Pruned.

\subsection{Adversarial Training and Attacks}
Adversarial robustness is introduced using Projected Gradient Descent based adversarial training. The total loss is defined as the mean of the clean loss and the adversarial loss. The clean loss is computed as the cross-entropy between the network’s predictions on unperturbed inputs and the corresponding ground-truth labels. The adversarial loss is computed as the cross-entropy between the predictions on adversarially perturbed inputs and the corresponding ground-truth labels. The final loss is expressed as
\begin{equation}
   L = \tfrac{1}{2} \left[ L_{\text{clean}} + L_{\text{adversarial}} \right]
\end{equation}
where both \(L_{\text{clean}}\) and \(L_{\text{adversarial}}\) are standard cross-entropy losses. Adversarial examples are generated during training using the network’s current parameters. The perturbation process follows the \(L-\infty\) Projected Gradient Descent method. Given a clean input \(X_{clean}\) and corresponding label \(y\), the input is iteratively perturbed in the direction of the sign of the loss gradient with respect to the input:
\begin{equation}
   X_{\text{adv}} = X_{\text{clean}} + \delta^{k}
\end{equation}
where \(\delta^{k}\) is the perturbation after \(k\) iterations of step size \(a\).
At each step, the perturbation is calculates as:
\begin{equation}
    \delta^{i+1} = \delta^{i} + a \cdot \text{sign}(\nabla_{x}L(\delta^{i}, y_{\text{true}}))
\end{equation}
After each update, the perturbation is clipped to ensure that it remains within the allowable \(\epsilon\)-bounded region around the clean input. Following this, a second clipping step confines the pixel intensities of the adversarial image to the valid range \([0,1]\). We use a step size \(a = 0.01\), number of iterations \(k = 40\) and epsilon \(\epsilon = 0.3\). Adversarial robustness at inference is evaluated by calculating both clean and adversarial accuracy under both FGSM and PGD attacks. The attacks are applied over a range of perturbation magnitudes, \(\epsilon\in[0, 0.5]\), with an increment of \(0.1\). Each configuration was tested across \(5\) independent runs with random initializations, and results were averaged.

\subsection{Hardware-Fault Simulation}
In this work, we emulate stuck-at-zero faults by progressively setting \(10\%\) of the remaining active weights to zero at each step, reaching up to \(80\%\) total faults. After each increment, clean and adversarial accuracies are evaluated to quantify the cumulative impact of hardware faults on network robustness.

\subsection{Evaluation Metrics}
Model robustness was assessed using clean accuracy, adversarial accuracy, with increasing fault levels. All experiments were repeated \(5\) times to ensure statistical reliability. Results are reported as mean \(\pm\) standard deviation across runs.
\begin{figure*}[!h]
    \centering
    \includegraphics[width=\textwidth]{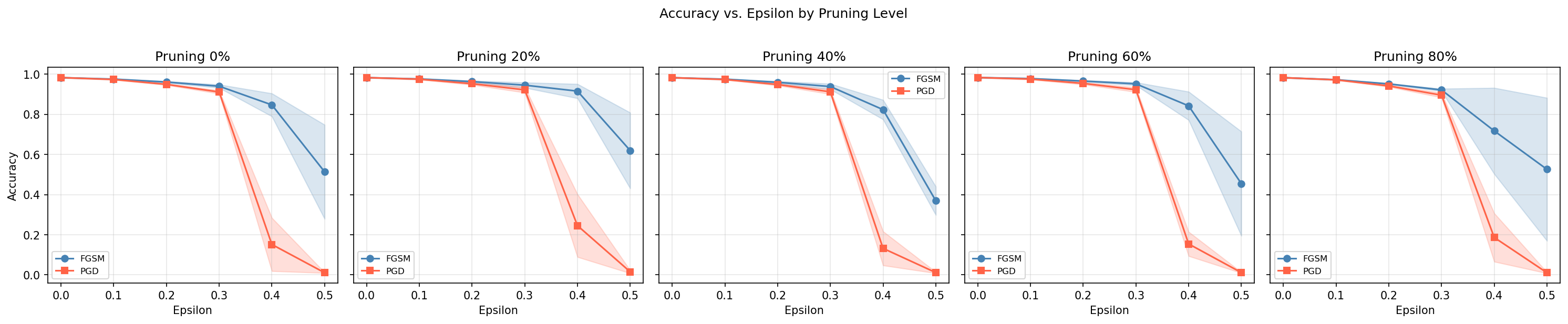}
    \caption{Accuracy vs. Attack Strength for Adversarially trained model at different pruning levels}
    \label{fig1}
\end{figure*}

\section{Results}
\subsection{Adversarial Training shows improved Input Robustness but reduces Fault Tolerance}
Figure~\ref{fig1} and Table~\ref{tab:results_full} show classification accuracy under FGSM and PGD attacks for the adversarially trained network across pruning levels. At $\epsilon=0.3$ and no faults present, FGSM and PGD accuracy remained above 89\% across all pruning levels, indicating that pruning does not significantly affect adversarial robustness under fault-free conditions.
\begin{figure}[!h]
    \centering
    \includegraphics[width=0.9\columnwidth]{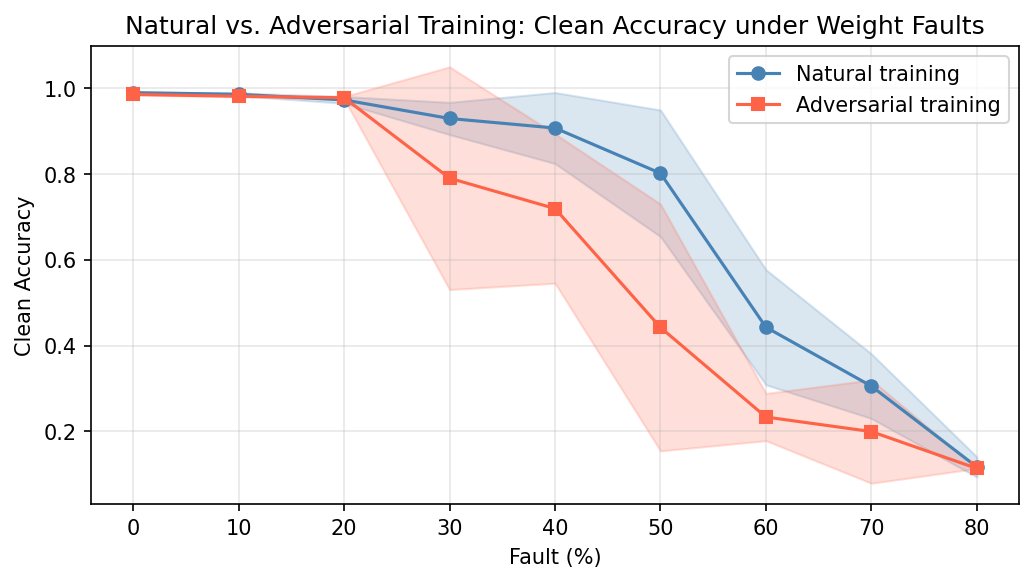}
    \caption{Clean Accuracy vs. Fault percentage for Natural training and Adversarial training}
    \label{fig2}
\end{figure}
\begin{table*}[t]
\centering
\caption{Clean and adversarial accuracy (mean $\pm$ std over 5 runs) at $\epsilon=0.3$, across pruning levels and stuck-at-zero fault rates.}
\label{tab:results_full}
\resizebox{\textwidth}{!}{%
\begin{tabular}{lccccccccc}
\toprule
\multirow{2}{*}{\textbf{Pruning}} & \multicolumn{3}{c}{\textbf{0\% Faults}} & \multicolumn{3}{c}{\textbf{40\% Faults}} & \multicolumn{3}{c}{\textbf{80\% Faults}} \\
\cmidrule(lr){2-4} \cmidrule(lr){5-7} \cmidrule(lr){8-10}
 & Clean & FGSM & PGD & Clean & FGSM & PGD & Clean & FGSM & PGD \\
\midrule
0\%  & 98.35$\pm$0.16 & 94.06$\pm$0.92 & 91.17$\pm$0.92 & 70.41$\pm$31.10 & 59.18$\pm$29.98 & 60.47$\pm$29.54 & 11.80$\pm$0.39 & 12.78$\pm$1.32 & 11.33$\pm$0.22 \\
20\% & 98.35$\pm$0.12 & 94.59$\pm$1.45 & 92.25$\pm$1.52 & 80.89$\pm$13.83 & 71.49$\pm$15.62 & 71.07$\pm$10.55 & 12.20$\pm$1.64 & 12.52$\pm$2.54 & 11.87$\pm$1.11 \\
40\% & 98.32$\pm$0.15 & 93.90$\pm$1.54 & 91.26$\pm$1.27 & 93.60$\pm$1.36 & 83.22$\pm$5.30 & 81.07$\pm$4.67 & 12.81$\pm$2.01 & 12.58$\pm$2.03 & 12.50$\pm$2.02 \\
60\% & 98.39$\pm$0.18 & 95.24$\pm$0.82 & 92.32$\pm$1.27 & 89.10$\pm$7.02 & 80.07$\pm$11.97 & 78.95$\pm$7.37 & 11.42$\pm$0.13 & 12.12$\pm$1.65 & 11.38$\pm$0.05 \\
80\% & 98.33$\pm$0.19 & 92.19$\pm$0.68 & 89.59$\pm$1.28 & 72.36$\pm$16.96 & 57.82$\pm$16.90 & 54.82$\pm$17.83 & 12.54$\pm$2.11 & 12.99$\pm$2.93 & 11.75$\pm$1.45 \\
\bottomrule
\end{tabular}%
}
\end{table*}
Figure~\ref{fig2} compares clean accuracy under increasing stuck-at-zero fault rates for naturally trained versus adversarially trained networks. Adversarially trained networks began degrading at lower fault rates than naturally trained networks, with the two curves diverging noticeably after 20\% faults. This observation is consistent with, but does not prove, our hypothesis that adversarial training shifts the decision boundary in a way that becomes increasingly fragile as weight faults accumulate.
\begin{figure*}[!h]
    \centering
    \includegraphics[width=\textwidth]{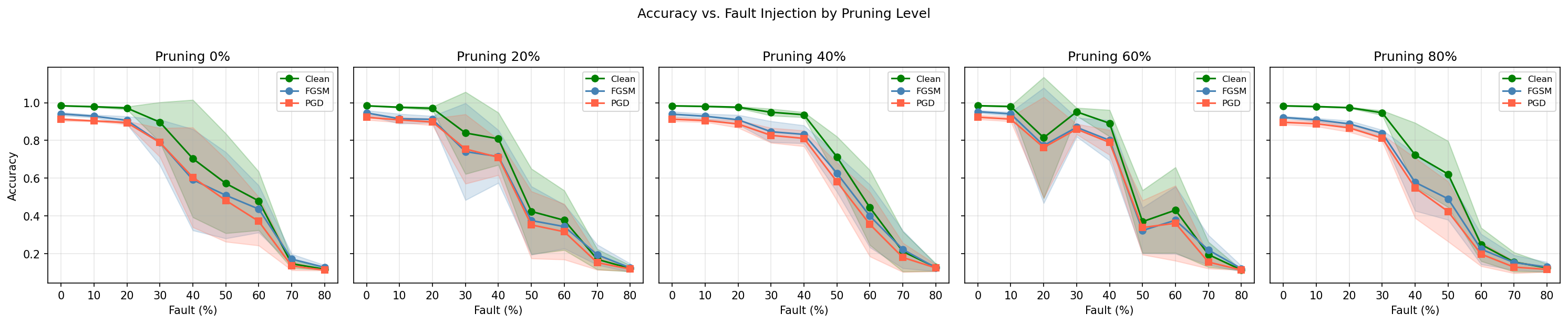}
    \caption{Accuracy vs. Fault percentage for Adversarially trained network at different pruning levels}
    \label{fig3}
\end{figure*}
\subsection{No Significant Increase in Fault Sensitivity was seen with Pruning}
Table~\ref{tab:results_full} shows accuracy at $\epsilon=0.3$ across pruning levels at three fault rates: 0\%, 40\%, and 80\%. At 0\% faults, clean accuracy remained stable across all pruning levels, showing that retraining after pruning fully recovers clean accuracy. Adversarial accuracy is lower than clean accuracy, as expected, but remains fairly stable across all pruning levels. At 40\% faults, accuracy varied substantially by pruning level, following a non-monotonic pattern. 40\% and 60\% pruning showed the highest retained clean and adversarial accuracy, while 0\% and 80\% pruning degraded more. At 80\% faults, all pruning levels collapsed to near chance level, with no meaningful separation between pruning levels. Figure~\ref{fig3} shows this non-monotonic pattern across the full 0-80\% fault range: clean, FGSM, and PGD accuracy degrade at a similar pace across all five pruning levels, with no pruning level showing a clear or consistent advantage. We note that variance at intermediate fault rates (20-40\%) was substantial, in some cases exceeding 30 percentage points of standard deviation across the 5 runs. This indicates that model behavior under moderate fault injection is highly sensitive to the specific random pattern of faulted weights. As a result, mean-accuracy comparisons alone may not support strong claims of statistical significance at these fault rates without further testing.

\subsection{Joint Characterization Across Fault Rate, Attack Strength, and Pruning}
\begin{figure*}[!h]
    \centering
    \includegraphics[width=\textwidth]{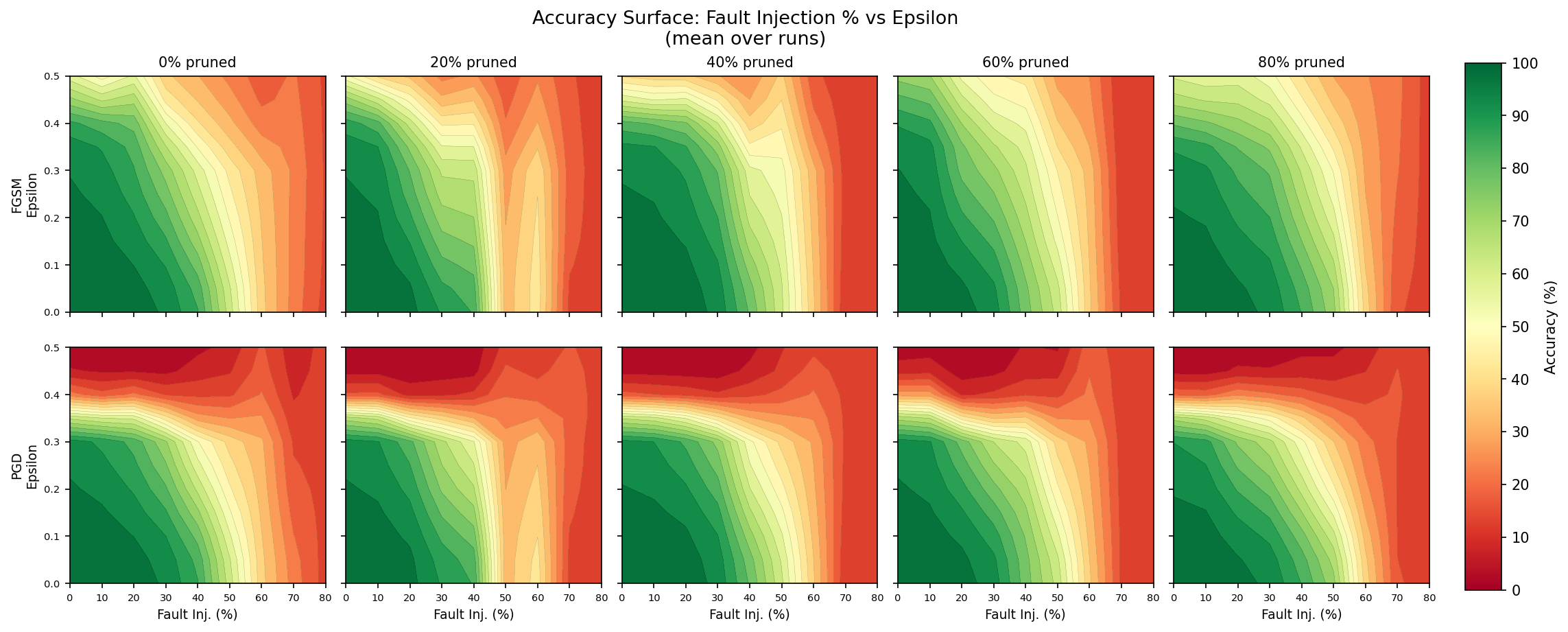}
    \caption{Joint characterization across fault rates, attack strengths, and pruning levels for Adversarially trained network}
    \label{fig4}
\end{figure*}

Figure~\ref{fig4} presents accuracy as a function of both fault injection rate and attack strength, for FGSM and PGD, across all five pruning levels. The overall shape of the accuracy surface is broadly consistent across pruning levels, showing that pruning level has limited impact on the joint robustness-fault trade-off. However, given the high variance noted above, this consistency reflects a general trend rather than a precise, reproducible boundary.

\section{Discussion}
The results show that adversarial training improves robustness against FGSM and PGD attacks but also increases sensitivity to hardware-induced weight faults (Figures~\ref{fig1}, \ref{fig2}). As the stuck-at-zero fault rate increases, adversarially trained models experience degradation in classification accuracy compared to naturally trained models (Figure~\ref{fig2}). Thus, the results reveal a non-trivial tradeoff between robustness to adversarial inputs and robustness to hardware perturbations. We propose an explanation grounded in decision boundary geometry. Adversarial training pushes the decision boundary further from the training data points, requiring larger perturbations to cause misclassifications. This creates what we consider an "optimal" boundary for defending against input attacks. However, when this learned network is exposed to stuck-at-zero weight faults, the corrupted weights alter the decision boundary geometry, resulting in a "sub-optimal" boundary. As faults accumulate, the boundary gets pushed back toward the data points. This means smaller adversarial perturbations are now sufficient to cause misclassifications, explaining the increased vulnerability of adversarially trained models under hardware faults. This is consistent with, though does not confirm, our hypothesis that adversarial training shifts the decision boundary in a way that becomes increasingly fragile as weight faults accumulate.

The joint characterization across fault rates, attack strengths, and pruning levels reveals that pruning level has minimal impact on the robustness-fault trade-off (Figure~\ref{fig4}). Magnitude-based pruning removes the smallest magnitude weights, keeping only the most significant weights that contribute most to the decision boundary.
We would intuitively expect adversarially trained pruned networks to be more prone to weight faults since they have fewer weights to represent information and faults would now corrupt these concentrated high importance weights. However, this is not what we observed. Adversarially trained pruned networks under the same percentage of weight faults show no additional sensitivity compared to their unpruned counterparts. The mechanism underlying this robustness warrants further investigation. A per-layer analysis comparing the magnitude distribution of faulted weights in pruned versus unpruned models would shed more light on this mechanism, which we leave for future work.
We also note that model behavior was highly variable at intermediate fault rates (20-40\%), with standard deviations across our five seeds sometimes exceeding 30 percentage points (Table~\ref{tab:results_full}). This indicates that the specific random pattern of faulted weights, not just the fault rate, can substantially affect a given model's degradation. This variability, combined with our limited sample size of five runs, means the trends we report here should be interpreted as consistent directional patterns rather than precisely characterized effects.
Moreover, this work focuses only on stuck-at-zero faults. Stuck-at-one faults, which may be more damaging in certain classification settings, represent an important complementary fault model for future study.
Future work will also include testing on additional datasets and model architectures, as well as validation on real neuromorphic hardware, to assess whether these trends generalize beyond the three-layer CNN and MNIST setting used in this study.
\section{Conclusion}
This paper presented a controlled experiment on how pruning, adversarial training, and hardware-induced weight faults jointly affect the reliability of convolutional neural networks. Using a three-layer CNN trained on MNIST, we found that adversarial training improves robustness against input perturbations but increases sensitivity to stuck-at-zero weight faults, which we explain through the geometry of the decision boundary. Contrary to intuition, pruning did not significantly increase sensitivity to these weight faults. Our joint characterization across fault rates, attack strengths, and pruning levels showed that pruning level has limited systematic impact on this trade-off, although the high variance observed across runs at intermediate fault rates suggests these trends should be interpreted directionally rather than as precise effects. These findings represent an initial step toward understanding the joint reliability of compressed, adversarially robust neural networks under hardware faults, with additional datasets, architectures, and fault types remaining important directions for future work.
\section*{AI Usage Statement}
Generative AI tools were used to assist with editing and polishing the manuscript, as well as supporting code development and debugging. All AI-assisted content was reviewed and verified by the authors, who take full responsibility for the final work.
\section*{Acknowledgments}
This material is based upon work supported by the Air Force Office of Scientific Research (AFOSR) under award FA9550-24-1-0322. Any opinions, findings, conclusions, or recommendations are those of the authors and do not necessarily reflect the views of the United States Air Force.
\printbibliography
\end{document}